\pdfoutput=1

\PassOptionsToPackage{table}{xcolor}

\documentclass[11pt]{article}

\usepackage[final]{acl}

\usepackage{times}
\usepackage{latexsym}

\usepackage[T1]{fontenc}

\usepackage[utf8]{inputenc}

\usepackage{microtype}

\usepackage{inconsolata}

\usepackage{graphicx}

\usepackage{amsmath,amssymb,amsfonts}
\usepackage{algorithmic}
\usepackage{textcomp}
\usepackage{subcaption}
\usepackage{tabularx}
\usepackage{color}
\usepackage{algorithm}
\usepackage{multirow}
\usepackage{xspace}
\usepackage{soul}
\usepackage{adjustbox}
\usepackage{enumitem}

\usepackage{wrapfig, lipsum, booktabs}
\usepackage{newtxtext} 
\usepackage{pgfplots}
\usetikzlibrary{pgfplots.groupplots}
\pgfplotsset{compat=1.3}
\usepackage{tikz}
\usetikzlibrary{patterns}
\usepackage{pgf-pie}

\newdimen\abovecrulesep
\newdimen\belowcrulesep
\makeatletter
\patchcmd{\@@@cmidrule}{\aboverulesep}{\abovecrulesep}{}{}
\patchcmd{\@xcmidrule}{\belowrulesep}{\belowcrulesep}{}{}

\definecolor{demphcolor}{RGB}{144, 144, 144}
\definecolor{mygray}{gray}{0.4}
\definecolor{lightgray}{rgb}{0.9, 0.9, 0.9}

\usepackage{adjustbox}
\newcommand{\impro}[1]{{\hspace{0.05cm}{\textbf{\textcolor{jweigreen}{\small(+#1)}}}}}

\definecolor{tabhighlight}{HTML}{e5e5e5}
\definecolor{grey}{RGB}{128,138,135}

\definecolor{oorange}{RGB}{215,122,71}
\definecolor{yyellow}{RGB}{230,185,79}
\definecolor{ppurple}{RGB}{122,30,97}
\definecolor{ggreen}{RGB}{112,173,71}

\definecolor{battleshipgrey}{rgb}{0.3, 0.3, 0.3}
\definecolor{brilliantrose}{rgb}{1.0, 0.33, 0.64}
\definecolor{americanrose}{rgb}{1.0, 0.01, 0.24}
\definecolor{jweigreen}{rgb}{0,0.45,0.24}
\definecolor{bluegray}{rgb}{0.1, 0.1, 0.4}
\definecolor{ao(english)}{rgb}{0.0, 0.5, 0.0}
\definecolor{blanchedalmond}{rgb}{1.0, 0.92, 0.8}
\definecolor{atomictangerine}{rgb}{1.0, 0.6, 0.4}
\definecolor{chocolate(web)}{rgb}{0.82, 0.41, 0.12}
\definecolor{bananayellow}{rgb}{1.0, 0.88, 0.21}
\definecolor{goldenbrown}{rgb}{0.6, 0.4, 0.08}
\definecolor{aliceblue}{rgb}{0.94, 0.97, 1.0}
\definecolor{beige}{rgb}{0.96, 0.96, 0.86}
\definecolor{babyblue}{rgb}{0.54, 0.81, 0.94}
\definecolor{camel}{rgb}{0.76, 0.6, 0.42}
\definecolor{cinnamon}{rgb}{0.82, 0.41, 0.12}

\definecolor{redlinkcolor}{rgb}{0.79607843, 0.25098039, 0.25882353}
\definecolor{bluecitecolor}{rgb}{0,0.36,0.69}
\usepackage{hyperref}
\usepackage{booktabs}
\usepackage{makecell}
\usepackage{multirow}
\usepackage{adjustbox}
\usepackage{xcolor}
\usepackage{tcolorbox}
\usepackage{subcaption}  
\usepackage{colortbl}
\usepackage{adjustbox}
\usepackage{makecell}
\usepackage{graphicx}
\usepackage{booktabs}
\usepackage{xspace}
\usepackage{fontawesome}  
\usepackage[bottom]{footmisc}  
\usepackage{bigfoot} 
\usepackage{colortbl}
\usepackage{tcolorbox}

\tcbuselibrary{breakable}

\newcommand{\model}{GenS\xspace}
\newcommand{\dataset}{GenS-Video-150K\xspace}
\newcommand{\videollms}{VideoLLMs\xspace}
\newcommand{\videollm}{VideoLLM\xspace}

\title{Generative Frame Sampler for Long Video Understanding}

\author{
\textbf{Linli Yao\textsuperscript{1}},
\textbf{Haoning Wu\textsuperscript{2}},
\textbf{Kun Ouyang\textsuperscript{1}},
\textbf{Yuanxing Zhang\textsuperscript{2}}, \\
\textbf{Caiming Xiong\textsuperscript{3}}, 
\textbf{Bei Chen\textsuperscript{4}},
\textbf{Xu Sun\textsuperscript{1}\thanks{Corresponding authors.}},
\textbf{Junnan Li\textsuperscript{3}\footnotemark[1]},
\\
\textsuperscript{1}National Key Laboratory for Multimedia Information Processing, \\ School of Computer Science, Peking University\\
 \textsuperscript{2}Peking University,
 \textsuperscript{3}Salesforce Research,
 \textsuperscript{4}Independent Researcher
\\
\texttt{\{linliyao,kunouyang\}@stu.pku.edu.cn} \quad
\texttt{\{longo,xusun\}@pku.edu.cn} \\
\texttt{\{realtimothyhwu,beibei1019\}@gmail.com} \quad
\texttt{\{cxiong,junnan.li\}@salesforce.com}
}
\begin{document}
\maketitle


\begin{figure*}[t!]
    \centering
    \includegraphics[width=0.96\textwidth]{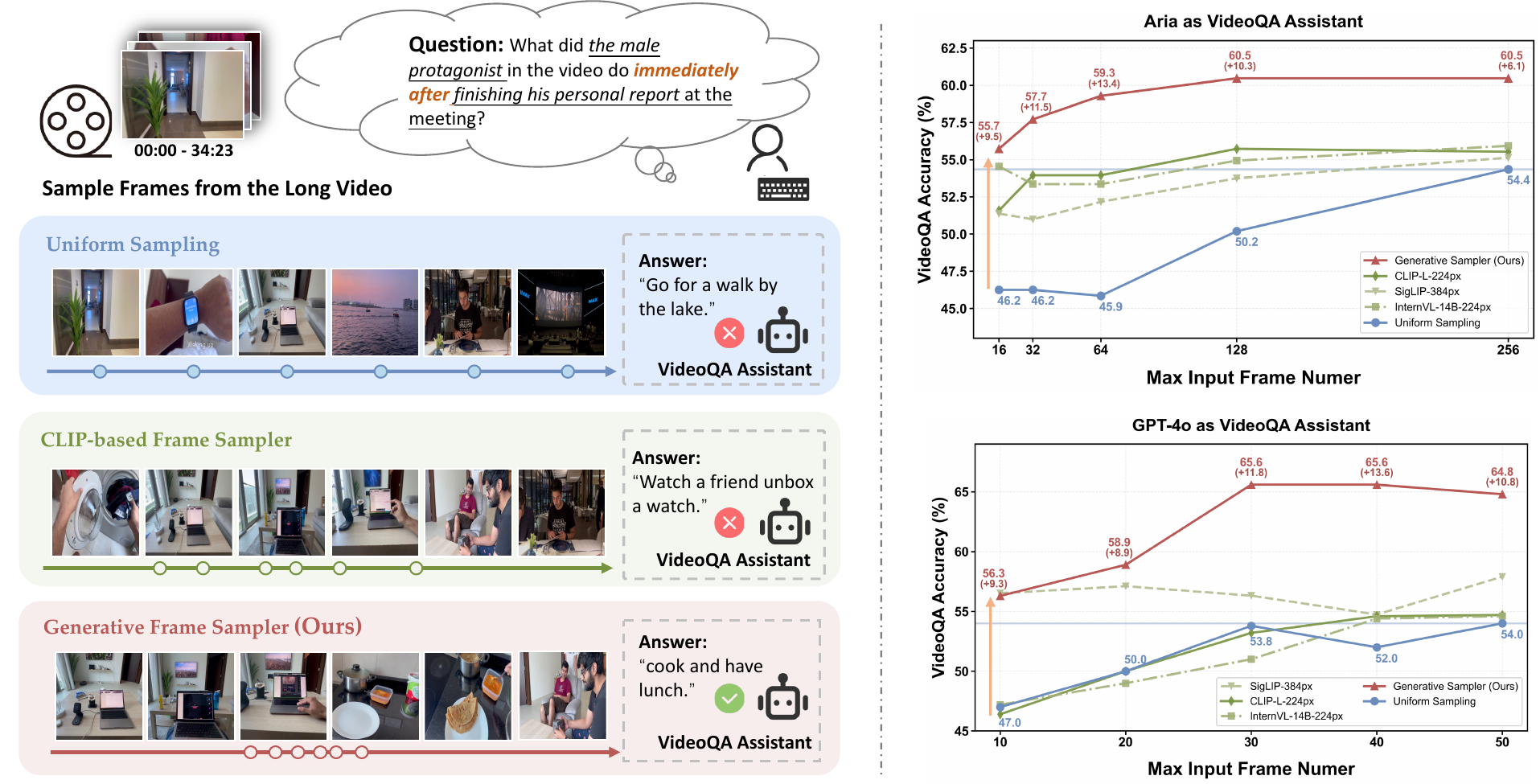}
    \caption{(a) An example of long video question-answering (VideoQA) using different frame samplers. Our Generative Frame Sampler (\model) accurately identifies relevant frame sequences based on the user question, further enhancing the performance of the downstream VideoQA assistant. (b) VideoQA accuracy results of state-of-the-art VideoQA assistants (Aria~\cite{liAriaOpenMultimodal2025} and GPT-4o~\cite{gpt4o}) when equipped with different frame samplers on the \textit{Vision-Centric subset} of LongVideoBench~\cite{longvideobench} .}
    \label{fig:intro}
\end{figure*}

\begin{abstract}
Despite recent advances in Video Large Language Models (\videollms), effectively understanding long-form videos remains a significant challenge. Perceiving lengthy videos containing thousands of frames poses substantial computational burden.
To mitigate this issue, this paper introduces \textbf{Gen}erative Frame \textbf{S}ampler (\textbf{\model}), a plug-and-play module integrated with \videollms to facilitate efficient lengthy video perception.
Built upon a lightweight \videollm, \model leverages its inherent vision-language capabilities to identify question-relevant frames.
To facilitate effective retrieval, we construct \textbf{\dataset}, a large-scale video instruction dataset with dense frame relevance annotations. Extensive experiments demonstrate that \model consistently boosts the performance of various \videollms, including open-source models (Qwen2-VL-7B, Aria-25B, VILA-40B, LLaVA-Video-7B/72B) and proprietary assistants (GPT-4o, Gemini). 
When equipped with \model, open-source \videollms achieve impressive state-of-the-art results on long-form video benchmarks: LLaVA-Video-72B reaches 66.8 (+4.3) on LongVideoBench and 77.0 (+2.7) on MLVU, while Aria obtains 39.2 on HourVideo surpassing the Gemini-1.5-pro by 1.9 points.
We release the code, dataset and models at \url{https://generative-sampler.github.io}.

\end{abstract}

\section{Introduction}

Recent advances in Large Multimodal Models (LMMs)~\cite{li2023blip2,dai2023instructblip, Qwen2VL, liu2023llava15,liu2024llavanext,chen2023internvl,tongcambrian} have shown remarkable progress, yet understanding long videos remains a significant challenge~\cite{zohar2024apollo,li2024temporal}. Current video-oriented LMMs~\citep{llava-video-sft,damonlpsg2023videollama,2023videochat,liAriaOpenMultimodal2025,zhang2024longva}, known as \videollms, typically employ an image encoder such as CLIP~\cite{clip} or SigLIP~\cite{siglip} to encode individual video frames as an initial step in video perception.
When processing hours-long videos containing thousands of frames, a critical challenge emerges: \textit{how to efficiently sample representative frames from the original video sequence?}

    

Existing \videollm assistants primarily employ two approaches for sampling lengthy videos: 1) \textit{uniform sampling} based on the \videollm's maximum context length, 
which leads to significant visual information loss due to limited fixed-interval sampling; 2) \textit{frame-per-second (FPS) sampling}, as implemented in long-context models like  Gemini~\citep{gemini}, which can capture frames at 1 FPS for comprehensive visual coverage. However, it obtains thousands of frames for hours-long videos, resulting in booming memory consumption and slow inference speed. 

Intuitively, for \videollm assistants, most frames in long videos are redundant when addressing a specific user instruction (i.e., \textit{query}).
To mitigate visual redundancy, several works propose language-guided frame sampling via CLIP to retrieve query-aware frames efficiently~\cite{arefeen2024vita,videoagent,mmniah}. 
However, CLIP-based frame samplers have three major limitations. For visual side, its frame-by-frame matching fails to capture temporal relationships implied by successive frames, as depicts in Figure~\ref{fig:intro} (a). For textual side, it is constrained by limited language capabilities, only able to process concise and simple user queries. Additionally, it embeds frames and textual queries separately to calculate cosine similarity, which hinders sufficient vision-language interaction to achieve complex multi-hop reasoning~\cite{videomme, longvideobench, hourvideo}.


To mitigate these limitations, we present \textbf{Gen}erative Frame \textbf{S}ampler (\textbf{\model}), a \videollm-based approach to retrieve relevant frames through flexible user instructions.
Built upon an advanced long-context \videollm~\citep{liAriaOpenMultimodal2025}, our approach inherits fundamental video-language perception capabilities.
\textbf{First}, as Figure~\ref{fig:intro} (a) illustrates, \model effectively captures temporal relationships between successive frames, such as \textit{`` \underline{immediately after}''}. \textbf{Second}, powered by built-in LLMs~\cite{llama3,vicuna2023}, \model comprehends complex and flexible textual instructions. \textbf{Third}, its native multi-modal architecture enables complex multi-hop reasoning by aligning long-range temporal cues with language semantics.
As demonstrated in Figure~\ref{fig:intro} (b), by selecting more relevant visual frames, \model substantially enhances the performance of VideoQA Assistants across both open-source (Aria~\cite{liAriaOpenMultimodal2025}) and proprietary (GPT-4o~\cite{gpt4o}) models. Compared with uniform sampling, \model improves Aria's accuracy by 13.4 points ($\leq$64 frames) and GPT-4o's accuracy by 13.6 points ($\leq$40 frames) on the challenging long-form video benchmark~\cite{longvideobench}. These significant improvements highlight that efficient video perception is a critical bottleneck for modern VideoQA Assistants, and \model provides an practical solution to boost their full potential.

To develop the \text{\model} sampler, we address two primary challenges:
Firstly, there is a \textit{shortage of training data}, as existing video instruction datasets~\cite{llava-video-sft,Maaz2023VideoChatGPT, etbench} lack dense annotations of relevant frames across diverse videos and user instructions.
Secondly, the optimal \textit{generative format for relevant frames sampling} remains under-explored.
To address the first challenge, we introduce \textbf{\dataset}, a novel synthetic VideoQA dataset with question-relevant frame annotations via GPT-4o.
The relevant frame annotations are: 1) \textit{dense}, with 20\% of all frames annotated, and 2) \textit{fine-grained}, with specific confidence scores (levels 1 to 5) assigned to each relevant frame.
For the second challenge, we explore different generative formats for indexing relevant frames. Empirical results show that directly appending textual labels (\texttt{``Frame Number [N]''}) before visual frames is sufficient to distinguish sequential frames. \model outputs the relevant frame spans with confidence scores as a natural language generation task (\texttt{\{``Frame $\text{N}_{start}$-$\text{N}_{end}$: relevance score'', ...\}}).

To summarize, our \textbf{main contributions} are threefold:
1) We propose \model, a novel generative frame sampler that leverages \videollms to identify question-aware relevant frames. It serves as a plug-and-play sampler that enhances input frames for VideoQA Assistants.
2) We introduce \dataset, a large-scale video instruction dataset that densely annotates relevant frames with fine-grained confidence scores across diverse video questions.
3) Through extensive experiments, we demonstrate that \model significantly enhances the performance of both open-source (Qwen2VL, Aria, VILA-v1.5, LLaVA-Video) and proprietary (GPT-4o and Gemini-1.5-pro) VideoQA Assistants. Notably, when equipped with \model, LLaVA-Video-72B achieves state-of-the-art performance with accuracy scores of 77.0 on MLVU and 66.8 on LongVideoBench, while Gemini1.5-pro attains 40.7 on HourVideo with averaging 45.7 minutes video duration.

\section{Method}
We introduce the novel \model method that effectively selects instruction-aware frames from long-form videos. 
To address the challenge of insufficient training data, we first construct \dataset, a video instruction dataset with dense relevant frame annotations (Section~\ref{sec:dataset}). We then present the \model architecture, focusing on an efficient generative format for VideoLLM-based frame retrieval (Section~\ref{sec:model}). Finally, we demonstrate how to integrate \model with existing VideoQA Assistants to enhance long-form video perception (Section~\ref{sec:frame_sampling}).

\subsection{\dataset Dataset Collection}
\label{sec:dataset}
Our objective is to construct \textit{(video, user instruction, relevant frames)} samples that enable the \model to identify salient frames for user instructions.
Existing datasets for grounded VideoQA~\cite{groundedvqa} and event localization~\cite{didemo,timechat,etbench,numpro} are limited by their \textit{domain specificity}, \textit{naive instruction}, and \textit{sparse key frame annotations}, make them inadequate for training robust frame samplers in real-world long-form video understanding. 
To address these limitations, we introduce \dataset across diverse video topics and flexible user instructions, with two key features: 1) \textit{dense} frame relevance annotations, with approximately 20\% of frames marked as relevant, and 2) \textit{fine-grained} scoring, where each relevant frame is assigned specific confidence scores (1-5).

We observe that even powerful proprietary LMMs like GPT-4o~\cite{gpt4o} struggle to achieve satisfactory retrieval performance when directly processing thousands of frames from lengthy videos (verified in Table~\ref{tab:baseline}).
To ensure high dataset quality, we decompose the synthetic data creation into a carefully designed four-stage pipeline leveraging GPT-4o. All prompts are provided in the Appendix~\ref{sec:supp_dataset}.
\paragraph{Stage 1: Dense Video Frame Captioning.}
We first curate a diverse collection of videos from YT-Temporal-1B~\cite{ytt1b}, encompassing a broad range of topics from YouTube~\footnote{https://www.youtube.com/}. Inspired by prior works~\cite{chen2024sharegpt4video}, we generate differential paragraph captions for each frame at a dense sampling rate (0.2 fps), focusing on distinguishing new visual content from previous frames. This dense frame captioning approach has been widely adopted as a preliminary step in video instruction dataset construction~\cite{chen2024sharegpt4video, llava-video-sft}.

\paragraph{Stage 2: Construct Video QAs with Grounded Frames.}
In this stage, we generate 12 distinct types of video question-answer (QA) pairs with grounded frames based on the dense frame captions. 
Specifically, we prompt GPT-4o to analyze every 50 consecutive frame captions and generate QA pairs of assigned types. Frames referenced during QA generation are marked as grounded frames (detailed in Appendix~\ref{sec:supp_dataset}).
To ensure robust generalization, we maintain a balanced distribution between generative and multiple-choice questions (50\% each). For multiple-choice questions, we augment the retrieval query by incorporating candidate options alongside the user question. We also strategically include 1\% negative samples (questions with no relevant frames) to enhance model robustness against irrelevant queries.

\paragraph{Stage 3: Extend Relevant Frames.}
We expand the set of relevant frames beyond those strictly grounded obtained in Stage 2.
GPT-4o typically references only a small subset of frames during VideoQA generation, resulting in a \textcolor{black}{\textit{low retrieval ratio} $R_f=\frac{N_{grd}}{N_{total}}$}, where $N_{grd}$ represents the number of grounded frames and $N_{total}$ denotes the total number of captioned frames.
Training with such a small $R_f$ would limit \model's ability to provide comprehensive frame coverage for long-context video understanding. Therefore, we employ CLIP-based retrieval to increase $R_f$ from \textcolor{black}{5\%} to 30\% approximately to add candidate relevant frames.

\paragraph{Stage 4: Score Fine-grained Relevant Confidence.}
Finally, we score the relevance (from 0 to 5, 0 is non-relevant and 5 is the most relevant) among all candidate relevant frames given the video and user question. This stage accurately refines the relevant frames in a global view of the whole video. Moreover, it provides fine-grained supervision to distinguish multi-level relevances to enable a top-K retrieval by confidence score duration inference.

\paragraph{Data Statistics.}
Our four-stage pipeline yields 150K data samples, with representative examples shown in Appendix Figure~\ref{fig:supp_data_case}. Table~\ref{tab:dataset_stats} provides a comprehensive overview of our dataset statistics. Each video has an average duration of 647.5 seconds (10.8 minutes) and contains approximately 129.5 captioned frames. Of these frames, an average of 20\% are annotated as relevant with fine-grained confidence scores ranging from 0 to 5 (where 0 indicates non-relevant and 5 indicates most relevant), thereby providing dense supervision.

\begin{table}[t]
\centering
\caption{Statistics of \dataset dataset.}
\label{tab:dataset_stats}
\begin{adjustbox}{max width=0.98\columnwidth}
\begin{tabular}{ll}
\toprule
\textbf{Property} & \textbf{Value} \\
\midrule
Format & \{video, question, answer, scored relevant frames\} \\
Total Samples & 150K \\
Avg Video Duration & 647.5 seconds (10.8 minutes) \\
Avg Task Number & 12 \\
Relevant Frame Rate & around 20\% \\
Relevance Scores & 0-5 (0: non-relevant, 5: most relevant) \\
\bottomrule
\end{tabular}

\end{adjustbox}
\end{table}

\subsection{\text{\model} Architecture}
\label{sec:model}
We design the \text{\model} architecture based on the state-of-the-art Aria model~\citep{liAriaOpenMultimodal2025}, which offers three key advantages:
1) As a native multimodal model, Aria demonstrates superior capability in understanding interleaved video-language contexts, enabling effective identification of relevant frames with textual indexing.
2) With a context window supporting up to 256 input frames, Aria's architecture excels at modeling temporal relationships implied in diverse user instructions.
3) Through its Mixture-of-Experts (MoE) architecture~\cite{moe,krajewski2024scalingmoe} with 3.9B activated parameters, Aria strikes an optimal balance between inference efficiency and multimodal performance compared to conventional 7B-parameter \videollms~\cite{llava-video-sft,Qwen2VL,chen2023internvl,lin2023vila}.

\subsubsection{Efficient Frame Indexing}
Leveraging Aria's advanced capabilities in encoding interleaved visual-textual representations, we implement an efficient frame indexing mechanism by prepending each frame with a textual number \texttt{[N]} that denotes the \texttt{[N-th Frame]}. This enables \model to uniquely identify and retrieve relevant frames based on their temporal positions.

For output representation, we adopt a JSON-based format similar to proprietary video assistants like Gemini and GPT-4o, where \model generates frame relevance predictions as a language modeling task. The output schema flexibly accommodates both discrete frame annotations (e.g., \texttt{\{``frame number'': relevance score\}}) and continuous temporal spans (e.g., \texttt{\{``start frame - end frame'': relevance score\}}) based on the retrieval context.
Our experiments in Section~\ref{sec:abla_fmt} demonstrate that organizing retrieved frames by relevance scores yields better performance compared to temporal ordering.

\subsubsection{Adaptation for Various Input FPS}
\model is designed as a plug-and-play frame sampling module that seamlessly integrates with existing VideoQA Assistants~\cite{llava-video-sft,Qwen2VL,chen2023internvl,lin2023vila, videoagent}. To handle varying candidate frame number and sampling densities across downstream VideoQA models, we implement a flexible frame retrieval mechanism that supports both dense (high FPS) and sparse (fixed-interval) frame sampling patterns. Specifically, we normalize the candidate frame indices to a unified range of 1-256 within each retrieval temporal window, ensuring robust retrieval performance regardless of the original frame sampling rate.

\subsection{Training and Inference Paradigm}
\label{sec:frame_sampling}
We train \model on both our \text{\dataset} and existing human-annotated event-level video datasets, specifically the E.T. Instruct dataset~\citep{etbench}, to enhance training data diversity. 
However, directly mixing E.T. Instruct data degrades \model's performance due to its sparse grounded frame annotations. Therefore, we integrate and post-process the E.T. Instruct dataset to better align with our frame sampling task (details in Section~\ref{sec:training_dataset}).

During inference, \model processes videos at arbitrary frame rates, retrieving instruction-relevant frames with confidence scores within each temporal window (maximum 256 frames). 
The retrieval across multiple temporal windows can be parallelized for efficient processing.
Output relevant frames are naturally sorted by confidence scores, with the number of relevant frames $N_{ret}$ varying based on the specific question and video content. We select the top $K$ frames for input to a VideoQA model, where $K = \min(N_{ret}, N_{ctx})$, with $N_{ctx}$ being the VideoQA model's maximum context length.

\begin{table*}[t]
    \small
    \centering
    \begin{adjustbox}{max width=1.97\columnwidth}
    \begin{tabular}{llccccc}
    \toprule
    \multirow{2}{*}{\textbf{VideoQA Model}} & \multirow{2}{*}{\textbf{Size}} & \multirow{2}{*}{\textbf{Sampled  Frames}} & \multicolumn{2}{c}{\textbf{LongVideoBench$_{\text{val}}$}~\textcolor{gray}{(avg 12min)}} & \multicolumn{2}{c}{\textbf{MLVU$_{\text{val}}$}~\textcolor{gray}{(avg 12min)}} \\
    \addlinespace[2pt]
    \cmidrule[0.5pt](lr){4-5} \cmidrule[0.5pt](lr){6-7}
    \addlinespace[2pt]
    & & & \textit{Full} & \textit{V-Centric} & \textit{Full} & \textit{V-Centric } \\
    \midrule 
    \rowcolor{aliceblue}\multicolumn{7}{l}{\textit{Proprietary LMMs}} \\
    GPT-4o & - & 256/0.5fps & 66.7 & - & 64.6 & - \\
    Gemini-1.5-Pro & - & 256/256 & 64.0 & - & - & - \\
    \midrule
    \rowcolor{aliceblue}\multicolumn{7}{l}{\textit{Open-source Video LLMs}} \\
   LLaVA-Video & 7B & 64/64 & 58.9 & 50.0 & 70.4 & 66.9 \\
    \rowcolor{gray!20}$\text{LLaVA-Video }_{\textbf{w/ \model}}$ & 7B & 54/50 &\phantom{00000}63.3~\textcolor{jweigreen}{ \small(+4.4)}  & \phantom{00000}56.7~\textcolor{jweigreen}{ \small(+6.7)} & \phantom{00000}73.4~\textcolor{jweigreen}{ \small(+3.0)} & \phantom{00000}70.6~\textcolor{jweigreen}{ \small(+3.7)} \\
    Qwen2-VL & 7B & 64/64 & 56.0 &45.9  & 64.7 & 62.3 \\
    \rowcolor{gray!20}$\text{Qwen2-VL }_{\textbf{w/ \model}}$ & 7B & 54/50 &   \phantom{00000}58.7~\textcolor{jweigreen}{ \small(+2.7)} & \phantom{00000}49.2~\textcolor{jweigreen}{ \small(+3.3)} & \phantom{00000}66.9~\textcolor{jweigreen}{ \small(+2.2)} & \phantom{00000}64.8~\textcolor{jweigreen}{ \small(+2.5)} \\
    
    Aria & 25B-A3.9B & 256/256 & 62.7 & 54.4 & 69.5 & 62.1 \\
    \rowcolor{gray!20}$\text{Aria }_{\textbf{w/ \model}}$ & 25B-A3.9B & 54/95 & \phantom{00000}66.1~\textcolor{jweigreen}{ \small(+3.4)} & \phantom{00000}\textbf{59.3}~\textcolor{jweigreen}{ \small(+4.9)} & \phantom{00000}72.6~\textcolor{jweigreen}{\small(+3.1)} & \phantom{00000}67.5~\textcolor{jweigreen}{\small(+5.4)} \\
     VILA-v1.5& 40B & 14/14 &57.4  &47.0  &57.8  &52.5 \\
    \rowcolor{gray!20}$\text{VILA-v1.5 }_{\textbf{w/ \model}}$ & 40B & 14/14 &   \phantom{00000}59.6~\textcolor{jweigreen}{ \small(+2.2)} & \phantom{00000}50.2~\textcolor{jweigreen}{ \small(+3.2)} & \phantom{00000}63.5~\textcolor{jweigreen}{ \small(+5.7)} & \phantom{00000}58.3~\textcolor{jweigreen}{ \small(+5.8)} \\
    
    LLaVA-Video & 72B & 64/64 & 62.5 & 51.6& 74.3 & 72.5 \\
    \rowcolor{gray!20}$\text{LLaVA-Video }_{\textbf{w/ \model}}$ & 72B & 54/50 &\phantom{00000}\textbf{66.8}~\textcolor{jweigreen}{ \small(+4.3)}  & \phantom{00000}58.9~\textcolor{jweigreen}{ \small(+7.3)} & \phantom{00000}\textbf{77.0}~\textcolor{jweigreen}{ \small(+2.7)} & \phantom{00000}\textbf{74.1}~\textcolor{jweigreen}{ \small(+1.6)} \\
    \bottomrule
    \end{tabular}
    \end{adjustbox}
    \caption{Performance on LongVideoBench~\citep{longvideobench} and MLVU~\citep{mlvu} benchmarks using multiple-choice accuracy metrics. \textit{V-Centric} denotes a vision-centric subset containing questions that explicitly require video understanding rather than language-only reasoning, while filtering short videos. Sampled Frames \textit{N/M} indicates sampled N frames for LongVideoBench and M frames for MLVU separately. 
    Using \model, we select the K most relevant frames (K <= max frame number of VideoQA models) and report the average number of input frames.}
    \label{tab:main_lvb}

\end{table*}

\begin{table*}[t]
    \small
    \centering
    \begin{adjustbox}{center, width=0.89\textwidth}

    \setlength\tabcolsep{2pt}
    \setlength\extrarowheight{3pt}
    \begin{tabular}{l|ccc|cccc|ccccccccc|cc|c}
    \hline
    & \multicolumn{3}{c}{\cellcolor[HTML]{CFE2F3}\textbf{Summarization}}
    & \multicolumn{4}{c}{\cellcolor[HTML]{B6D7A8}\textbf{Perception}}
    & \multicolumn{9}{c}{\cellcolor[HTML]{D9D2E9}\textbf{Visual Reasoning}}
    & \multicolumn{2}{c}{\cellcolor[HTML]{FFE599}\textbf{Navigation}} &
    \textbf{Avg} \\
    \hline
    &\rotatebox{90}{Key Events/ Objects} &\rotatebox{90}{Temporal Sequencing } &\rotatebox{90}{Compare/ Contrast} &\rotatebox{90}{Factual Recall} &\rotatebox{90}{Sequence Recall} & 
    \rotatebox{90}{Temporal Distance}
    &\rotatebox{90}{Tracking} &\rotatebox{90}{Relationship} &\rotatebox{90}{Proximity} 
    &
    \rotatebox{90}{Layout} &
    \rotatebox{90}{Duration} &\rotatebox{90}{Frequency} &\rotatebox{90}{Pre-requisites} &\rotatebox{90}{Predictive} &\rotatebox{90}{Causal} &\rotatebox{90}{Counterfactual} &\rotatebox{90}{Room-to-Room} &\rotatebox{90}{Object Retrieval} \\ \hline
    \cellcolor{aliceblue}Blind LLMs & & & & & & & & & & & & & & & & & & \\
    GPT-4 &22.7 &29.6 &24.2 &21.9 &15.2 &20.6 &15.8 &14.9 &21.4 &22.2 &23.6 &19.3 &14.7 &14.5 &18.7 &21.2 &15.8 &18.8 &19.6 \\
    \hline
    \cellcolor{aliceblue}Socratic Models & & & & & & & & & & & & & & & & & & \\
    LLaVA-34B-DPO &34.0 &35.5 &35.8 &30.3 &19.3 &12.7 &34.5 &18.3 &15.3 &26.7 &21.3 &17.9 &23.5 &20.9 &21.3 &22.4 &20.8 &22.4 &22.3 \\
    GPT-4 &40.5 &41.5 &43.2 &33.1 &20.0 &20.2 &36.7 &18.5 &21.7 &37.8 &25.3 &22.9 &27.1 &24.1 &24.7 &26.5 &20.0 &26.6 &25.7 \\
    \hline
    \cellcolor{aliceblue}Multimodal Models & & & & & & & & & & & & & & & & & & \\
    \text{Aria} \textcolor{grey}{(256 frm.)} &\text{58.2} &53.9 &\text{55.8} &44.7 &33.8 &\text{28.1} &41.9 &26.8 &36.9 &\text{28.9} &\text{42.3} &34.9 &50.0 &54.8 &31.3 &23.8 &15.0 &25.0 &38.7 \\
    \rowcolor{gray!20}\text{Aria\textsubscript{ \text{w/ \text{\model}}} \textcolor{grey}{(226 frm.)}} &56.3 &54.6 &50.5 &44.5 &\text{34.7} &26.6 &\text{45.3} &26.8 &38.3 &26.7 &\text{42.7} &36.0 &\text{50.8} &\text{56.8} &33.3 &29.1 &15.8 &22.9 &39.2 \\
    
    Gemini1.5-Pro \textcolor{grey}{(0.5fps)} &56.4 &\text{59.5} &46.7 &41.8 &33.6 &19.7 &35.7 &27.4 &38.2 &21.4 &37.2 &35.4 &46.8 &46.3 &41.0 &\text{38.7} &19.2 &33.9 &37.3 \\

    \rowcolor{gray!20}Gemini1.5-Pro\textsubscript{ \text{w/ \text{\model}}}  &57.6 & 57.9 & 53.7 & 45.3 & 34.7 & 26.2 & 45.8 & 31.7 & 39.2 & 15.6 & 39.8 & 40.8 & 48.9 & 49.4 & 48.7 & 37.1 & 29.2 & 34.9 & \textbf{40.7} \\
    \rowcolor{gray!20}\textcolor{grey}{(344 frm.)} & \textcolor{grey}{+1.2} & \textcolor{grey}{-1.6} & \textcolor{grey}{+7.0} & \textcolor{grey}{+3.5} & \textcolor{grey}{+1.1} & \textcolor{grey}{+6.5} & \textcolor{grey}{+10.1} & \textcolor{grey}{+4.3} & \textcolor{grey}{+1.0} & \textcolor{grey}{-5.8} & \textcolor{grey}{+2.6} & \textcolor{grey}{+5.4} & \textcolor{grey}{+2.1} & \textcolor{grey}{+3.1} & \textcolor{grey}{+7.7} & \textcolor{grey}{-1.6} & \textcolor{grey}{+10.0} & \textcolor{grey}{+1.0} & \textcolor{red}{+3.4} \\

    \bottomrule
    \end{tabular}
    \end{adjustbox}
    \caption{Results on HourVideo~\citep{hourvideo} benchmark, an extremely challenging video dataset with an average duration of 45.7 minutes, containing 113 videos longer than 60 minutes. \textit{Blind LLMs} perform reasoning without video inputs. \textit{Socratic models} first segment videos into one-minute intervals, generate captions for each segment using LLaVA-34B-DPO or GPT-4, then use GPT-4 to answer questions based on the aggregated captions. \textit{Multimodal Models} directly process video inputs for inference. }
    \label{tab:hourvideo}
\end{table*}

\section{Experiments}

\subsection{Experimental Settings}
\textbf{Evaluation Benchmarks.}
\label{sec:benchmarks}
We evaluate \model on several long-form video benchmarks including $\text{LongVideoBench}_\text{val}$ (LVB)~\cite{longvideobench}, $\text{MLVU}_{\text{Dev}}$~\cite{mlvu}, and HourVideo~\cite{hourvideo}. These benchmarks assess multiple-choice question-answering accuracy on videos ranging from minutes to hours in duration. 
For LVB and MLVU, we construct a more challenging \textit{Vision-Centric subset} by filtering out both questions answerable through pure language reasoning and videos of short duration.
Additionally, we evaluate the zero-shot temporal grounding capability of \model on the Charades-STA~\cite{charades-sta} dataset using mean Intersection over Union (mIoU) and Recall@1 at IoU thresholds of 0.3, 0.5, and 0.7.

\noindent\textbf{Training Dataset.}
\label{sec:training_dataset}
We utilize timestamp-output tasks from E.T.Instruct 164K~\cite{etbench}, extracting 75K base training samples (denoted as E.T. Instruct-75K). To enhance the density of grounded frame annotations, we post-process these samples through timestamp label aggregation and textual query concatenation within each video, yielding 41K samples (denoted as $\text{E.T. Instruct-41K}_{\text{agg.}}$). 
The final training data for \model combines this aggregated E.T. Instruct dataset with \dataset.

\noindent\textbf{Implementation Details.}
We train the open-source Aria~\footnote{https://github.com/rhymes-ai/Aria} model with a frozen vision encoder. The model supports a maximum sequence length of 32K tokens, accommodating up to 256 frames per sequence. Training consists of 300 iterations with a global batch size of 256, completed in 10 hours using 32 H800 GPUs. We provide complete hyperparameter settings in Appendix~\ref{sec:supp_hyperpara}.
During inference, the MOE architecture of \model utilizes only 3.9B activated parameters. We obtain original frames from the input video at 1 FPS to ensure comprehensive visual coverage, and then sample frames within each 256-frame interval using a sliding window approach. 
For multiple-choice questions, we append candidate options to the retrieval query.

\subsection{Results on Long-form Video Tasks}
\model functions as a plug-and-play frame sampling module that enhances visual perception capabilities across VideoQA models. We evaluate its effectiveness across three categories of VideoQA models:
I) \textit{Advanced proprietary \videollms}, specifically GPT-4o~\cite{gpt4o} and Gemini-1.5-pro~\cite{gemini};
II) \textit{Open-source competitive \videollms} with standard context lengths (64 input frames), including LLaVA-Video-7B/72B~\cite{llava-video-sft} and Qwen2-VL-7B~\cite{Qwen2VL};
III) \textit{Open-source long-context \videollms}, represented by Aria-25B~\cite{liAriaOpenMultimodal2025} with 256-frame input capacity. 

For MLVU and LongVideoBench, we construct a more challenging \textit{Vision-Centric subset} (also filtering out short videos) based on two observations: 1) several questions in the original datasets can be answered through pure language reasoning without visual context; 2) visual content in short videos can be adequately captured through uniform frame sampling. Details are provided in Appendix~\ref{sec:supp_hyperpara}.

\vspace{-0.1cm}
\paragraph{LongVideoBench.} 
As shown in Table~\ref{tab:main_lvb}, \model demonstrates consistent improvements across different VideoQA models and sizes. For standard-context models (64 frames), \model enhances LLaVA-Video-7B and Qwen2-VL-7B by 4.4 and 2.7 points, respectively, on the full validation set. Notably, even for long-context models like Aria-25B (256 frames), \model still brings a significant 3.4-point improvement, highlighting the importance of efficient frame sampling beyond model context length scaling. When equipped with \model, LLaVA-Video-72B achieves 66.8\% accuracy on LongVideoBench, establishing a new state-of-the-art. These gains become more pronounced on the \textit{Vision-Centric subset}, where \model improves LLaVA-Video-72B by 7.3 points and Aria-25B by 4.9 points, demonstrating its particular effectiveness on questions that demand stronger visual understanding capabilities.
Figure~\ref{fig:intro} (b) indicates that \model also significantly improves the performance of GPT-4o, achieving a 13.6\% accuracy gain with 40 input frames.

\vspace{-5pt}

\paragraph{MLVU.} 
On the MLVU benchmark, \model consistently enhances the performance of various VideoQA models. LLaVA-Video-7B's accuracy improves by 3.0 points (from 70.4\% to 73.4\%), Qwen2-VL-7B shows a 2.2-point increase (from 64.7\% to 66.9\%), and Aria demonstrates a 3.1-point gain (from 69.5\% to 72.6\%). Most notably, LLaVA-Video-72B integrated with \model achieves state-of-the-art performance with 77.0\% accuracy.

\vspace{-5pt}
\paragraph{HourVideo.} 
We further evaluate \model on HourVideo (Table~\ref{tab:hourvideo}), a particularly challenging benchmark featuring videos with an average duration of 45.7 minutes, including 113 videos that exceed one hour in length. Prior to our work, only Gemini-1.5-Pro could process such extensive videos end-to-end, achieving 37.3\% accuracy. With the integration of \model, both Aria-25B and Gemini-1.5-Pro surpass previous results, reaching 39.2\% and 40.7\% accuracy respectively, thereby establishing new state-of-the-art performance. These improvements demonstrate \model's capability to effectively process extremely long videos through its dynamic frame identification.

\subsection{Comparison with Sampling Baselines}

We evaluate \model against various frame sampling approaches in Table~\ref{tab:baseline}. Specifically, we compare against uniform sampling baseline to assess the effectiveness of these methods.

\textit{Image-language matching methods} like CLIP-L-224px~\cite{clip} and SigLIP-384px~\cite{siglip} demonstrate significant improvements over uniform sampling only with sparse frame inputs (e.g., 16 frames), as shown in \textcolor{black}{Figure~\ref{fig:intro} (b)}. However, when processing 256 frames, they achieve only slight gains (0.7/1.1 points) over uniform sampling, due to their limitations in complex language reasoning and temporal relationship modeling.
As Table~\ref{tab:baseline} shows, InternVL-14B-224px~\cite{mmniah} outperform CLIP-based approaches, benefiting from their enhanced language understanding capabilities.

\textit{Advanced Proprietary LMMs} like GPT-4o unexpectedly yield only a 1.3-point improvement over 256-frame uniform sampling. Our empirical observations reveal that GPT-4o struggles with precise frame selection when processing large candidate frame sets, particularly in identifying correct frame indices. 
While multi-round refinement or step-by-step reasoning could potentially address these limitations, it would be prohibitively expensive.

\textit{\videollm-based Methods} like FRAME-VOYAGER~\cite{framevoyager} specialize in sampling sparse frames from a candidate pool (e.g., 8 from 128 frames). Following their experimental settings, \model also shows effectiveness for enhancing short-context VideoQA models such as VILA-1.5-40B (14 frames)~\cite{lin2023vila}, surpassing FRAME-VOYAGER by 2.4 points and uniform sampling by 5.7 points. Event localization methods such as TimeChat~\cite{timechat} can identify event timestamps based on textual descriptions, but show only marginal improvements over uniform sampling when used for frame sampling, likely due to the coarse and sparse nature of event localization annotations.

\begin{table}[t]
    \small
    \centering
    \begin{adjustbox}{max width=0.95\columnwidth}
    \begin{tabular}{lc}
    \toprule
    \multirow{2}{*}{} & \multicolumn{1}{c}{\textbf{LongVideoBench$_{\text{val}}$}} \\
    & \textit{(V-Centric Subset)} \\
    \midrule
    \textbf{Aria-25B as VideoQA Model (<=256 frames)} \\
    \hline
    \addlinespace[2pt]
    +Uniform Sampling  & 54.4\\
    \addlinespace[2pt]
    \rowcolor{aliceblue}\multicolumn{2}{l}{\textit{Image-Language Matching}} \\
    \addlinespace[2pt]
    +CLIP-L-224px Sampler & 55.5 \\
    +SigLIP-384px Sampler & 55.1 \\
    +InternVL-14B-224px Sampler & 55.9 \\
    \addlinespace[2pt]
    \rowcolor{aliceblue}\multicolumn{2}{l}{\textit{Proprietary LMMs}} \\
    \addlinespace[2pt]
    +GPT-4o Sampler & 55.7\\
    \addlinespace[2pt]
    \rowcolor{aliceblue}\multicolumn{2}{l}{\textit{Open-source VideoLLMs}} \\
    \addlinespace[2pt]
    \rowcolor{gray!20}+\text{\model} (ours)  & \textbf{60.5}\\
    \midrule
    \midrule
    \addlinespace[2pt]
     & \textbf{MLVU\textsubscript{M-avg}} \\
    \midrule
    \textbf{VILA-v1.5-40B as VideoQA Model (<=14 frames)}\\
    \hline
    \addlinespace[2pt]
    +Uniform Sampling & 57.8 \\
    \addlinespace[2pt]
    \rowcolor{aliceblue}\multicolumn{2}{l}{\textit{Specialized VideoLLMs for Event Localization}} \\
    \addlinespace[2pt]
    +TimeChat Sampler\textsubscript{\textcolor{grey}{[CVPR 2024]}} & 59.4\\
    \addlinespace[2pt]
    \rowcolor{aliceblue}\multicolumn{2}{l}{\textit{Specialized VideoLLMs for Frame Sampling}} \\
    \addlinespace[2pt]
    +FRAME-VOYAGER Sampler\textsubscript{\textcolor{grey}{[ICLR 2025]}} & 61.1 \\
    \rowcolor{gray!20}+\text{\model} (ours) & \textbf{63.5} \\
    
    \bottomrule
    \end{tabular}
    \end{adjustbox}
    \vspace{-5pt}
    \caption{Comparison with different frame sampling methods.
    }
    \label{tab:baseline}
    \vspace{-10pt}
\end{table}

\subsection{Results on Temporal Grounding Tasks}
Table~\ref{tab:charades} demonstrates that \model achieves competitive video grounding performance, surpassing GPT-4o and approaching specialized \videollms like TimeSuite~\cite{timesuite}. This highlights its excellence in both long-form video understanding and fine-grained temporal localization. Notably, our model achieves these results without training on any Charades-STA data. Even when excluding \textit{E.T.Instruct-41K$_{\text{agg.}}$} that contains temporal grounding data from DiDeMo (8.4K samples), Queryd (661 samples), and TACoS (61 samples)~\cite{didemo, queryd, tacos}, our model's performance remains comparable to GPT-4o.


\begin{table}[t]
    \small
    \centering
    \begin{adjustbox}{max width=\columnwidth}
    \begin{tabular}{lrrrr}
    \toprule
    \multirow{2}{*}{\textbf{Grounding Model}} & \multicolumn{4}{c}{\textbf{Charades-STA}} \\
    & $\text{R1@{0.3}}$ & $\text{R1@{0.5}}$ & $\text{R1@{0.7}}$ & $\text{mIoU}$ \\
    \midrule
    \rowcolor{aliceblue}\multicolumn{5}{l}{\textit{Temporal Grounding \videollms (7B size)}} \\
    VTimeLLM & 51.0 & 27.5 & 11.4 & 31.2 \\
    HawkEye & 50.6 & 31.4 & 14.5 & 33.7 \\
    TimeChat\textsubscript{\textcolor{grey}{[CVPR 2024]}}  & - & 32.2 & 13.4 & 30.6 \\
    TimeSuite\textsubscript{\textcolor{grey}{[ICLR 2025]}}  & 69.9 & 48.7 & 24.0 & - \\
    \addlinespace[2pt]
    \rowcolor{aliceblue}\multicolumn{5}{l}{\textit{General \videollms}} \\
    GPT-4o & 55.0 & 32.0 & 11.5 & 35.4 \\
    VideoChat2-7B & 9.6 & 3.4 & 1.4 & - \\
    Qwen2-VL-7B & 8.7 & 5.4 & 2.4 & 7.9 \\
    LongVA-7B-DPO & 22.6 & 10.1 & 2.2 & 14.6 \\
    LLaVA-OneVison-7B & 31.2 & 13.5 & 5.2 & - \\
    Aria & 39.0 & 18.6 & 6.6 & 26.7 \\
    \midrule
    \rowcolor{gray!20} \model & 62.9 & 38.7 & 15.2 & 38.0 \\
    $\text{\model}_{\textbf{~ \text{w/o E.T.Instruct-41K$_{\text{agg.}}$}}}$ & 51.1 & 28.2 & 10.4 & 33.2 \\
    \bottomrule
    \end{tabular}
    \end{adjustbox}
    \vspace{-5pt}
    \caption{Results on the Charades-STA~\cite{charades-sta} temporal grounding benchmark. }
    \label{tab:charades}
    \vspace{-5pt}
\end{table}

\begin{table*}[t]
    \small
    \centering
    \begin{adjustbox}{max width=0.98\textwidth}
    \begin{tabular}{lllllllll}
    \toprule
    & \multicolumn{2}{c}{\cellcolor{white}\textbf{Holistic}} 
    & \multicolumn{2}{c}{\cellcolor{white}\textbf{Multi Detail}} 
    & \multicolumn{3}{c}{\cellcolor{white}\textbf{Single Detail}} 
    & \cellcolor{white}\textbf{M-Avg} \\
    \cmidrule(lr){2-3}\cmidrule(lr){4-5}\cmidrule(lr){6-8}
    \addlinespace[1pt]
    & {Topic Reason} 
    & {Anomaly Recognition} 
    & {Action Order} 
    & {Action Count} 
    & {Needle QA} 
    & {Ego Reason} 
    & {PlotQA} 
    & {} \\
    \midrule
    Uniform & 87.1 & 69.5 & 63.7 & 44.2 & 77.8 & 66.2 & 71.8 & 69.9 \\
    \rowcolor{gray!10} \text{\model} & 84.7 \textcolor{grey!90}{(-2.4)} & 72.0 \textcolor{jweigreen}{(+2.5)} & 73.3 \textcolor{jweigreen}{(+9.6)} & 41.3 \textcolor{grey!90}{(-2.9)} & 85.4 \textcolor{jweigreen}{(+7.6)} & 66.8 \textcolor{jweigreen}{(+0.6)} & 76.4 \textcolor{jweigreen}{(+4.6)} & 73.2 \textcolor{jweigreen}{(+3.3)} \\
    \bottomrule
    \end{tabular}
    \end{adjustbox}
    \caption{Breakdown performance analysis on MLVU (val) using Aria as the VideoQA model with 256 input frames. }
    \label{tab:mlvu_break}
\end{table*}

\begin{table}[t]
    \small
    \centering
    \begin{adjustbox}{max width=0.95\columnwidth}
    \begin{tabular}{lcc}
    \toprule
    \multirow{2}{*}{\textbf{VideoQA Models}} & GPT-4o  & Aria \\
    & \text{(<=32 frm.)} & \text{(<=256 frm.)} \\
    \midrule
    \textbf{Uniform Sampling} & 53.4 & 54.4 \\
    \midrule
    \textbf{Frame Sampler}  &  &  \\
      + \dataset & 63.8 & 57.7 \\
    \addlinespace[2pt]
    \rowcolor{aliceblue}\multicolumn{3}{l}{\textit{Directly Mixing Dataset}} \\
     + \dataset + DiDeMo-40K & 61.5 & 57.7 \\
     + \dataset + E.T.Instruct-75K  & 62.7 & 59.5 \\
    \addlinespace[2pt]
    \rowcolor{aliceblue}\multicolumn{3}{l}{\textit{With Query Aggregation}} \\
     + \dataset + E.T.Instruct-41K$_{\text{agg.}}$ & 63.8 & \textbf{60.9} \\
    \textit{(unified task prompts)} & & \\
    \rowcolor{gray!20} + \dataset + E.T.Instruct-41K$_{\text{agg.}}$ & \textbf{64.2} & 60.5 \\
    \rowcolor{gray!20} \textit{(distinct task prompts)} & & \\
    \bottomrule
    \end{tabular}
    \end{adjustbox}
    \caption{Ablation study on different training datasets and combination strategies. Results are accuracy (\%) on LongVideoBench$_{\text{val}}$ \textit{(V-Centric Subset)}.}
    \label{tab:dataset}
    \vspace{-5pt}
\end{table}

\subsection{Analysis}
\paragraph{Effectiveness of \dataset Dataset.}

Table~\ref{tab:dataset} presents that adding our \dataset brings remarkable improvements over the uniform sampling baseline across two VideoQA models. For GPT-4o with 32 frames input, adding VC-RAG-150K improves accuracy by 10.4 points (from 53.4 to 63.8). For Aria with 256 frames input, the improvement is 3.3 points (from 54.4 to 57.7). 
We further evaluate the combination of our \dataset with temporal grounding data DiDeMo~\cite{didemo} and time-sensitive video dataset E.T.Instruct~\cite{etbench}. Experiments reveal that directly combining these datasets degrades GPT-4o's performance due to cross-task inconsistencies. However, applying query aggregation post-processing (described in Section~\ref{sec:training_dataset}) leads to improved overall performance.
We also compare two prompting strategies: using a single unified prompt as frame-sampling tasks versus using task-specific prompts. Our results show that task-specific prompts perform better, since they allow the model to learn specialized behaviors for each task type while still benefiting from the combined training data.

\paragraph{Input and Output Indexing Format.}
\label{sec:abla_fmt}

\begin{figure}[t]
    \centering
    \includegraphics[width=0.95\linewidth]{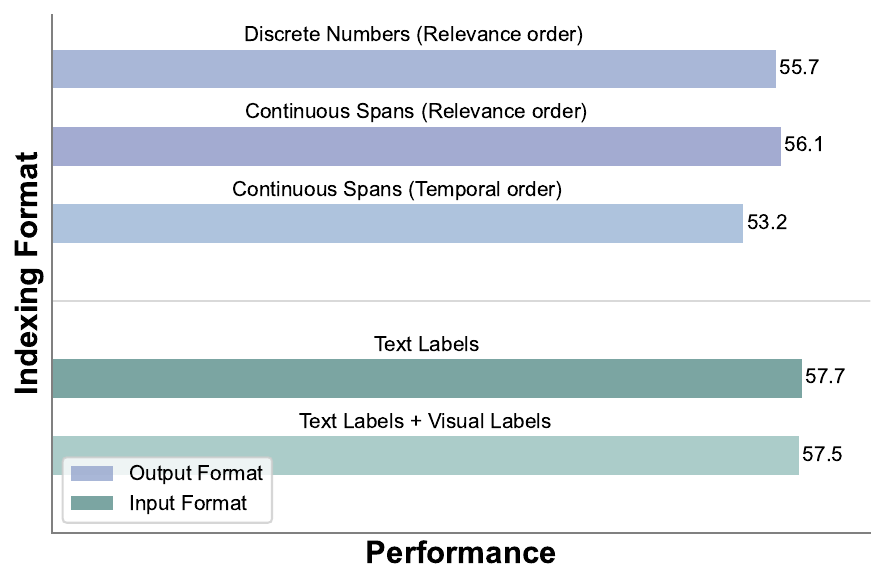}
    \caption{Ablation study on different input and output frame indexing formats.}
    \label{fig:abla_fmt}
    \vspace{-5pt}
\end{figure}

For output formats, we evaluate two key aspects: (1) use discrete index numbers versus integrate successive frames into continuous spans, and (2) order frames chronologically or by relevance. Figure~\ref{fig:abla_fmt} depicts that \textit{continuous spans} with confidence scores \textit{ordered by relevance} achieve the best performance (56.1).
For input formats, we compare two strategies: (1) \textit{textual indexing alone} that prepending each frame with a textual number \texttt{[N]} and (2) \textit{combining textual and visual indexing} which additionally overlays visual numerical indices directly onto each frame at the pixel level~\cite{wu2024numpro}.
Our results show that \textit{textual indexing alone} performs marginally better than \textit{combining textual and visual indexing}, indicating that \model can effectively process interleaved visual-textual sequences.

\paragraph{Breakdown results on different question types.}
Table~\ref{tab:mlvu_break} shows that \model achieves significant improvements on question types that require precise temporal understanding and localization. Specifically, \model brings notable gains on Needle QA (+7.6) and Action Order (+9.6) tasks, where identifying specific moments or temporal relationships between actions is crucial. 
However, for Topic Reason tasks that require holistic video understanding, uniform sampling provides better coverage of the overall video content.

\section{Extension Applications}

\noindent\textbf{Coarse-to-Fine Hybrid Sampling.}
We propose a coarse-to-fine hybrid approach that combines a lightweight CLIP~\cite{clip}  sampler with our \model to improve sampling efficiency for extremely long videos. Specifically, we first adopt CLIP to densely sample frames from the 1 fps candidate pool and return top 256 most relevant frames, then apply \model to re-sample the most informative frames within a single 256-frame temporal window. 

Table~\ref{tab:hybrid} demonstrates that this hybrid approach consistently outperforms both uniform sampling and standalone CLIP sampling across various frame count constraints. We also provide the \model sampled at 1fps as an upper bound for the frame sampling efficiency-performance tradeoff. This demonstrates that a simple hybrid approach can effectively improve the efficiency of the \model sampler while maintaining competitive performance, which provides a more practical solution for real-world applications.

\begin{table*}[t]
\centering

\begin{adjustbox}{max width=1.7\columnwidth}
\begin{tabular}{llllll}
\toprule
\textbf{\#Sampled Frames} & \textbf{$\leq$10 frm} & \textbf{$\leq$20 frm} & \textbf{$\leq$30 frm} & \textbf{$\leq$40 frm} & \textbf{$\leq$50 frm} \\
\midrule
Uniform & 47.04 & 50.00 & 53.75 & 51.98 & 53.95 \\
CLIP-L-224px & 46.44 & 50.00 & 53.16 & 54.55 & 54.74 \\
CLIP-L-224px + GenS & \textbf{53.16} \impro{6.12} & \textbf{56.13} \impro{6.13} & \textbf{60.47} \impro{6.72} & \textbf{61.26} \impro{6.71} & \textbf{60.86} \impro{6.12} \\
\midrule
\textcolor{grey}{GenS~(Upper Bound)}  & \textcolor{grey}{56.32} & \textcolor{grey}{58.89} & \textcolor{grey}{65.61} & \textcolor{grey}{65.61} & \textcolor{grey}{64.82} \\
\bottomrule
\end{tabular}
\end{adjustbox}
\caption{Performance of a hybrid sampling approach on LongVideoBench$_{\text{val}}$ \textit{(V-Centric Subset)} using GPT-4o as VideoQA model. The hybrid approach first retrieves the top 256 relevant frames from a 1 FPS candidate pool using CLIP-L-224px, then applies \model to select the most informative frames within a single 256-frame temporal window.}
\label{tab:hybrid}
\end{table*}

\begin{table*}[t]
    \centering
    
    \begin{adjustbox}{max width=1.7\columnwidth}
    \begin{tabular}{llllll}
    \toprule
    \textbf{\#Sampled Frames} & \textbf{$\leq$10 frm} & \textbf{$\leq$20 frm} & \textbf{$\leq$30 frm} & \textbf{$\leq$40 frm} & \textbf{$\leq$50 frm} \\
    \midrule
    Uniform & 47.04 & 50.00 & 53.75 & 51.98 & 53.95 \\
    InternVL-14B-224px & 47.23 & 49.01 & 50.99 & 54.35 & 54.55 \\
    CLIP-L-224px & 46.44 & 50.00 & 53.16 & 54.55 & 54.74 \\
    GenS\textsubscript{Qwen2.5VL-3B} & \textbf{54.74} \impro{7.51} & \textbf{55.93} \impro{5.93} & \textbf{57.11} \impro{3.36} & \textbf{59.68} \impro{5.13} & \textbf{58.69} \impro{3.95} \\
    \bottomrule
    \end{tabular}
    \end{adjustbox}
    \caption{Performance of \model with Qwen2.5VL-3B as the base \videollm using low-resolution input (112$\times$112 pixels) on LongVideoBench$_{\text{val}}$ \textit{(V-Centric Subset)}, with GPT-4o as the VideoQA model. }
    \label{tab:qwen_adapt}
    \end{table*}

\noindent\textbf{\model Implementation on Qwen2.5VL-3B.}
Our design of \textit{generative frame sampling} is not limited to a specific \videollm (e.g., Aria) as the base model. To verify the generalizability of our approach, we implemented \model on Qwen2.5VL-3B using low-resolution inputs (112$\times$112 pixels) for frame sampling.

Results in Table~\ref{tab:qwen_adapt} demonstrate that GenS based on Qwen2.5VL-3B achieves remarkable performance compared to both uniform sampling and CLIP-based samplers. The model shows consistent improvements across all frame count configurations, with gains of up to 7.51 points when using just 10 frames. 
The successful adaptation of \model to the distinctly different Qwen2.5VL-3B architecture validates the broad generalizability of our approach. Our method can be integrated with various advanced \videollms without requiring architectural modifications, enabling it to leverage ongoing advancements in the field.

\section{Related Work}
\subsection{Long-form Video Understanding}
Current video assistants~\cite{2023videochat,llava-video-sft,
lin2023vila,Qwen2VL,internvl1.5,li2024llava} have demonstrated impressive capabilities in video-language understanding. However, processing hours-long videos (e.g., 3600 frames per hour at 1 fps) for comprehensive visual coverage remains computationally prohibitive. Recent approaches address this limitation through either: 1) extending model context length to accommodate more frames (e.g., 256-512)~\cite{zhang2024longva, 
liu2024nvila, videoccam, wang2024longllava, wang2025internvideo2, 
longvila}, or 2) performing visual token compression within the model~\cite{li2024llamavid, deco, zhang2025videollama3,
videochat-falsh}. In contrast, we propose a more efficient paradigm - incorporating a frame sampler prior to model input, thus eliminating redundant visual processing inside the large-scale video assistants.

\subsection{\videollms with Retrieval-Augmented Generation}
To enhance video-language interaction, recent works have equipped \videollm assistants with Retrieval-Augmented Generation (RAG). Unlike text-based retrieval methods, i.e., Video-RAG~\cite{video-rag}, Q-ViD~\cite{romero2024question}, and R2A~\cite{Retrieving-to-answer}, we propose a visual-centric approach that directly retrieves relevant frames. Compared to CLIP-based retrieval~\cite{mmniah,arefeen2024vita,videoagent,EgoInstructor,tang2025adaptive}, our method built on a \videollm excels at capturing long-range temporal perception and complex language understanding. While similar frame samplers~\cite{framevoyager,mdp3} are limited to sparse sampling (e.g., 8 from 128 frames), our approach can efficiently retrieve thousands of frames with adaptive sampling rates, 
substantially enhancing long-context video assistants on hours-long video perception.

\section{Conclusion}
This paper presents \model, a novel generative frame sampling method and a high-quality video instruction dataset \dataset.
Our extensive experiments show that \model brings consistent improvements across different VideoQA models' architectures and sizes and achieve new state-of-the-art results on LongVideoBench (66.9), MLVU (77.0), and HourVideo (40.7). It suggests that efficient frame sampling is a promising direction for advancing long-form video understanding.

\section*{Limitations}
Leveraging \text{\model} for key frame retrieval in long-form videos incurs additional computational overhead compared to naive uniform sampling. Specifically, while uniform sampling processes N frames (where N is the context length of the video question-answering model), our approach needs to analyze M frames (M=256) within each retrieval window. However, this computational cost can be mitigated through parallel processing of multiple segment windows, making the overall inference time practically manageable.
Meanwhile, for large-scale advanced VideoQA Assistants like LLaVA-Video-72B, sampling few relevant frames via \model (3.9B activated parameters) is more efficient than substantially extending the model context length of a 72B VideoQA Assistant.
The performance of \text{\model} could be further enhanced through multi-round retrieval iterations and integration with Video Agent systems for refined frame selection.

\section*{Acknowledgements}
This research was partially supported by the National Natural Science Foundation of China under Grant No. 92470205 and No. 62176002. Xu Sun and Junnan Li are the corresponding authors.


\bibliographystyle{acl_natbib}
\bibliography{custom}

\clearpage
\appendix
\section{Appendix}
\label{sec:appendix}

\subsection{Additional Results and Analysis}
\label{sec:supp_results}

\paragraph{Leveraging Video Subtitles for Frame Sampling.}
As shown in Table~\ref{tab:subtitle_ret}, incorporating video subtitle information during frame sampling improves the quality of selected frames and enhances downstream VideoQA model performance. Specifically, across different input scales ranging from 16 to 256 frames, adding subtitle information yields consistent improvements of 0.6-2.77 points. This demonstrates \model's ability to effectively integrate visual and textual cues while maintaining efficient frame sampling. Notably, \model achieves this performance without explicit training on frame-subtitle sequences. This zero-shot capability stems from \model's base VideoLLM architecture, which inherently supports processing interleaved vision-text inputs.

\begin{table}[h]
    \small
    \centering
    \begin{tabular}{lccccc}  
    \toprule
    \multirow{2}{*}{\textbf{Sampling}} & \multicolumn{5}{c}{\textbf{Max Input Frames}} \\  
    \cmidrule(lr){2-6}
    \addlinespace[2pt]
    & $\leq16$ & $\leq32$ & $\leq64$ & $\leq128$ & $\leq256$ \\
    \midrule
    w/o subtitle & 54.74 & 54.74 & 55.34 & 55.53 & 54.74 \\
    w/ subtitle & 55.34 & 54.74 & 55.34 & 56.72 & 57.51 \\
    \midrule
    & +0.60 & +0.00 & +0.00 & +1.19 & +2.77 \\
    \bottomrule
    \end{tabular}
    \caption{Impact of incorporating video subtitles during frame sampling, evaluated on LongVideoBench (V-Centric) using Aria as the VideoQA model. The subtitles are only used by the frame sampler \model, not by the VideoQA model. Results show accuracy (\%) for different maximum frame budgets.}
    \label{tab:subtitle_ret}
    \end{table}

\paragraph{Frame Sampling Parameters during Inference.}
We investigate two key parameters in our frame sampling approach: sampling ratio and temporal window size. The sampling ratio determines how densely we sample candidate frames from the original video (measured in frames per second), while the temporal window size (temp\_win\_size) controls how many consecutive frames are considered simultaneously during sampling. As shown in Table~\ref{tab:sampling_param}, increasing the sampling ratio from 0.2 to 1.0 fps with temp\_win\_size=128 significantly improves performance from 52.37 to 55.13, as denser sampling provides more comprehensive video coverage. With the sampling ratio fixed at 1.0 fps, expanding the temporal window from 128 to 256 frames yields a further improvement to 56.13, demonstrating the benefit of longer-range temporal perception. Notably, during training, we use an average sampling ratio of 0.2 fps and temp\_win\_size of 129 frames. The superior performance achieved with different parameters during inference suggests that \model generalizes well beyond its training configuration rather than overfitting to the training settings.

\begin{table}[h]
    \small
    \centering
    \begin{tabular}{lc}
    \toprule
    \textbf{Sampling Configuration} & \textbf{Accuracy} \\
    \midrule
    sample\_ratio=0.2, temp\_win\_size=128 & 52.37 \\
    sample\_ratio=1.0, temp\_win\_size=128 & 55.13 \\
    sample\_ratio=1.0, temp\_win\_size=256 & \textbf{56.13} \\
    \bottomrule
    \end{tabular}
    \caption{Impact of sampling ratio and temporal window size on LongVideoBench (V-Centric). Higher sampling ratio enables denser frame coverage, while larger temporal window allows longer-range temporal perception.}
    \label{tab:sampling_param}
\end{table}

\subsection{\dataset Dataset Details}
\label{sec:supp_dataset}

\paragraph{Prompts for \dataset Annotation.}
We provide detailed prompts for each stage of the \dataset annotation process via GPT-4o. Table~\ref{tab:supp_prompt_stage1} depicts the prompts for \textbf{Stage 2 Construct Grounded Video QAs}, while Table~\ref{tab:supp_stage4_score} shows the prompts for \textbf{Stage 4 Score Frame Relevance}.

When constructing Video QAs with grounded frames (Stage 2), we define 12 specific question types to comprehensively cover different aspects of video understanding capabilities:
\begin{itemize}
    \item \textbf{Reasoning Tasks:} Object, Action, Spatial, and Temporal Reasoning questions test the model's ability to make logical inferences about relationships and changes in the video.
    \item \textbf{Perception Tasks:} Object, Action, Attribute, and Spatial Perception questions focus on basic visual understanding of scenes, actions, and object properties.
    \item \textbf{Specialized Tasks:} Video Detail Referring requires fine-grained visual attention, Counting tests quantitative understanding, OCR evaluates text recognition, and Temporal Perception assesses understanding of event sequences.
\end{itemize}


\subsection{Training Hyper-parameters and Evaluation Details}
\label{sec:supp_hyperpara}
\begin{table}
    \small
    \centering
    \begin{tabular}{lr}
    \toprule
    \textbf{Hyper-parameter} & \textbf{Value} \\
    \midrule
    \multicolumn{2}{l}{\textcolor{grey}{\textit{Visual Encoder}}} \\
    \midrule
    Frame Sampling Rate & Varied FPS (0.2-1.0) \\
    Input Resolution & 490 \\
    Visual Tokens per Image & 128 \\
    Max Image per Sequence & 256 \\
    Patch Size & 14x14 \\
    \midrule
    \multicolumn{2}{l}{\textcolor{grey}{\textit{Large Language Model (MOE)}}} \\
    \midrule
    Number of Layers & 28 \\
    Hidden Size & 2560 \\
    FFN Hidden Size & 13568 \\
    MOE FFN Dimension & 1664 \\
    Number of Attention Heads & 20 \\
    Number of KV Heads & 20 \\
    Number of Experts & 64 \\
    Top-k Experts & 6 \\
    Number of Shared Experts & 2 \\
    \midrule
    \multicolumn{2}{l}{\textcolor{grey}{\textit{Model Training}}} \\
    \midrule
    Max Context Length & 32768 \\
    Batch Size & 256 \\
    Learning Rate & 1e-5 \\
    Min Learning Rate & 1e-8 \\
    Warmup Ratio & 0.0 \\
    Training Iterations & 300 \\
    Z Loss & 1e-5 \\
    EP ST Load Balancing Loss & 1e-3 \\
    LR Scheduler Type & Cosine \\
    \bottomrule
    \end{tabular}
    \caption{Training hyper-parameters for \text{\model}.}
    \label{tab:supp_hyperpara}
\end{table}
We provide training hyper-parameters in Table~\ref{tab:supp_hyperpara}. 

\paragraph{Video-Centric Subset.} We use GPT-4o to filter questions that can be answered by purely textual reasoning on LongVideoBench (LVB) and MLVU. The filtered dataset contains 506 samples for LVB (excluding 159 non-vision-centric questions and 672 videos shorter than 10 minutes) and 879 samples for MLVU (excluding 200 non-vision-centric questions and 1,095 videos shorter than 8 minutes). All remaining questions in this subset explicitly require long-range visual understanding capabilities.

\subsection{Frame Sampling Baseline Implementation}
\paragraph{CLIP / SigLIP / InternVL.}
For image-language models like CLIP, SigLIP and InternVL, we implement frame sampling through similarity-based retrieval. We first densely sample frames from the original video at 1 FPS and extract visual features for each frame. We then encode the input question into text features and compute cosine similarity scores between each frame and the question embedding. Finally, we select the top-K frames with highest similarity scores as key frames, where K is determined by the maximum input frame capacity of the VideoQA model.

\paragraph{TimeChat.} 
For event localization \videollms like TimeChat, we use the question as a textual query to identify relevant event timestamps in the video. We then uniformly sample K frames from these identified temporal segments as key frames for downstream processing.

\paragraph{GPT-4o.}
We leverage GPT-4o's vision capabilities to score frame relevance based on the prompt template in Table~\ref{tab:supp_prompt_gpt4o_rag}. Since GPT-4o has limitations in processing massive frames at once, we first sample frames at 1 FPS from the original video and divide them into windows of 50 frames each. GPT-4o then processes each window independently to identify relevant frames. The relevant frames from all windows are aggregated to obtain K candidate frames. If K exceeds the VideoQA model's maximum input capacity N, we randomly sample N frames from the candidate set to form the final input.

\begin{table}[t]
    \centering
    \small
    
    \begin{tcolorbox}

\textbf{Prompt as a Frame Retrieval Assistant:}\\
You are an advanced AI visual assistant tasked with assessing frame relevance for question answering. Please retrieve the video frames relevant to the question (maybe with options) to answer it correctly, output the frame timestamp, exactly in format [XX:XX], [XX:XX, XX:XX], or [XX:XX, XX:XX, XX:XX], etc. If no matching frames are found, output [None].

\textbf{Video Frames:}
\begin{verbatim}
[00:00] <image_placeholder>
[00:05] <image_placeholder>
[00:10] <image_placeholder>
...
\end{verbatim}
    
\textbf{Question:} \begin{verbatim}<question_placeholder>\end{verbatim}\\
\textbf{Output Relevance Frames:}
\begin{verbatim}
[17:07, 17:26, 18:24]
\end{verbatim}
\end{tcolorbox}
\caption{The prompt used by GPT-4o to retrieve relevant video frames for question answering.}
\label{tab:supp_prompt_gpt4o_rag}
\end{table}

\begin{table*}[t]
    \centering
    \small
    
    \begin{tcolorbox}
\textbf{GPT-4 Prompt for Grounded Video Question Generation (Stage 2):}

You are a teacher designing challenging questions for a "Long-term Video Understanding" class. Your task is to create questions that test students' ability to comprehend and analyze long video content. I will provide video frame narrations with timestamps, and you should generate questions following the criteria below:\\

\textbf{Question Types:}

\begin{itemize}
    \item  \textbf{Specific requirements for a question type:} \begin{verbatim}<Question Type Placeholder>\end{verbatim}
    \item \textbf{Multiple Choice:} Create questions with five options (A-E), where only one is correct. All options should be closely related to the video content but clearly distinguishable.
    
    \item \textbf{Open-ended:} Create questions requiring concise, specific answers that can be directly supported by the video content.
\end{itemize}

\textbf{Key Requirements:}
\begin{enumerate}
    \item \textbf{Video-Centric:} Questions must be answerable solely through careful analysis of the video content. Avoid requiring external knowledge.
    
    \item \textbf{Temporal Reasoning:} Questions should require understanding relationships between events across different timestamps.
    
    \item \textbf{Clear Answers:} Ensure answers are concise, accurate, and directly supported by video evidence.
    
    \item \textbf{Difficulty:} Make questions challenging by requiring careful analysis of multiple video segments.
    
    \item \textbf{Format:} Do not reference scene indices or specific timestamps in questions.
\end{enumerate}

\textbf{Input Format:}
\begin{verbatim}
[Timestamp] Description of video frame
[04:30] Person walks into room
[04:35] Person picks up book
...
\end{verbatim}

\textbf{Output Format:} \\
For Multiple Choice Questions:
\begin{verbatim}
{
    "question": "...",
    "options": {
        "A": "...", "B": "...", "C": "...", "D": "...", "E": "..."
    },
    "correct_option": "A/B/C/D/E",
    "rationale_timestamps": ["04:30", "04:35", ...]
}
\end{verbatim}

For Open-ended Questions:
\begin{verbatim}
{
    "question": "...",
    "answer": "...",
    "rationale_timestamps": ["04:30", "04:35", ...]
}
\end{verbatim}

\textbf{Additional Guidelines:}
\begin{itemize}
    \item Make wrong options slightly longer than correct ones
    \item Distribute correct answers evenly across options A-E
    \item Include only timestamps directly relevant to the question
    \item Ensure answers compress information from multiple timestamps
\end{itemize}
    \end{tcolorbox}
    \caption{The prompt template used for \textbf{Stage 2 - Construct Grounded Video QAs} of \dataset.}
    \label{tab:supp_prompt_stage1}
\end{table*}
\begin{table*}[t]
    \centering
    \small
    
    \begin{tcolorbox}
\textbf{GPT-4o Prompt for Scoring Relevance (Stage 4):}

You are a helpful and precise assistant designed to evaluate the relevance of each video frame to a given textual question. I will provide video frames (or their narrations) along with their timestamps as input. Your task is to assign an overall relevance score on a scale from 1 to 5 for each video frame, where higher scores indicate better alignment with the question. Use the criteria below to guide your scoring: \\

\textbf{Scoring Criteria:}
\begin{itemize}
    \item \textbf{5 (Highly Relevant):} The video frame contains unique visual cues critical to accurately answering the question. Without this frame, it would be challenging to reason or provide an accurate answer.
    
    \item \textbf{4 (Directly Relevant):} The video frame is directly related to the question, but its visual cues can be partially replaced or supplemented by other important frames.
    
    \item \textbf{3 (Moderately Relevant):} The video frame is important for addressing the question, helping identify related scenes, activities, actions, individuals, or other elements, or aiding in ruling out incorrect options.
    
    \item \textbf{2 (Somewhat Relevant):} The video frame indirectly relates to the question, providing supporting context that aids in reasoning or finding the correct answer.
    
    \item \textbf{1 (Minimally Relevant):} The video frame has minimal relevance to the question. While it may involve the same person, activity, or action as the question, it does not contribute meaningfully to answering it.
    
    \item \textbf{0 (Irrelevant):} The video frame has no relevance to the question.
\end{itemize}

\textbf{Additional Notes:}
\begin{enumerate}
    \item Some textual questions may require multi-hop reasoning, necessitating the combination of visual cues from multiple frames to arrive at the correct answer.
    
    \item Some questions may ask about the global information of the video, such as identifying its main focus or summarizing the content. In these cases, assign higher scores to frames with unique and non-redundant visual information to ensure the selected frames collectively provide a comprehensive summary of the video while minimizing redundancy.
    
    \item Most input video frames will have some relevance to the question, so prioritize scoring between 1 and 5. Use a score of 0 only for entirely irrelevant frames.
    
    \item If the question is: (1) Ambiguous, such that none of the input frames can provide an answer, or (2) Contains logical issues (e.g., contradictions or nonsensical reasoning), then the question should be flagged as low quality, and the output should be "the question has low quality".
\end{enumerate}

\textbf{Input Frames:}
\begin{verbatim}
[04:40] <image_placeholder>
[04:45] <image_placeholder>
[04:50] <image_placeholder>
...
\end{verbatim}

\textbf{Question:} \begin{verbatim}<question_placeholder>\end{verbatim} \\

\textbf{Hint:}\\
Frames at timestamps \begin{verbatim}<timestamp_placeholder>\end{verbatim} are considered ``Highly Relevant'' and can be assigned a score of 5. \\

\textbf{Output Format:}\\
Provide an explanation for the assigned scores to justify your reasoning. Return the results in the following JSON format:
\begin{verbatim}
{
    "[04:40]": score 0-5,
    "[04:45]": score 0-5,
    ...
}
``Explain why each score was assigned, detailing the relevance of the frames to the question...''
\end{verbatim}
    \end{tcolorbox}
    \caption{The prompt template used for \textbf{Stage 4 - Score Frame Relevance} of \dataset.}
    \label{tab:supp_stage4_score}
\end{table*}

\subsection{Visualization Cases}
\begin{figure}[t]
    \centering
    \includegraphics[width=0.5\textwidth]{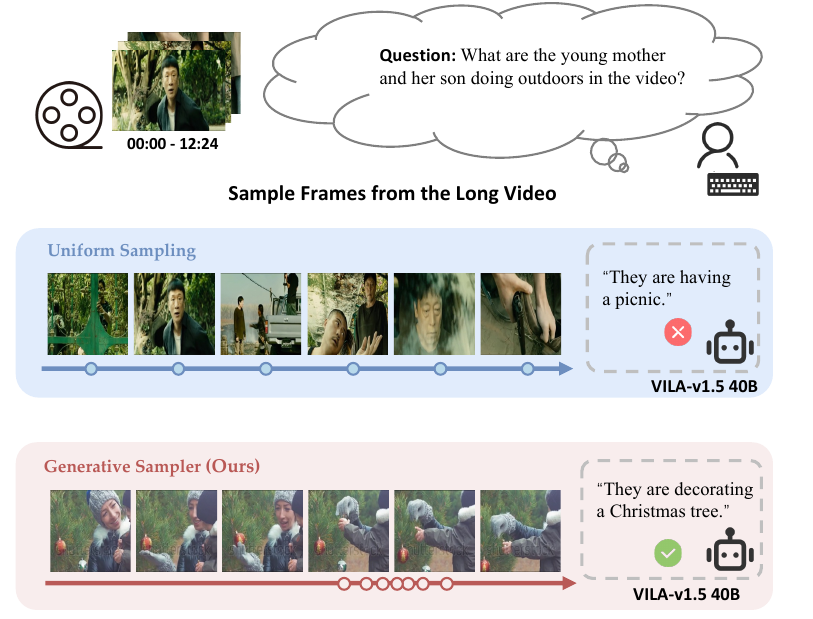}
    \caption{Visualization of \model integrated with VILA-v1.5-40B (<=14frames) on MLVU dataset.}
    \label{fig:supp_cases}
\end{figure}

We visualize a case in Figure~\ref{fig:supp_cases}.

\begin{figure*}[t]
    \centering
    \includegraphics[width=0.9\textwidth]{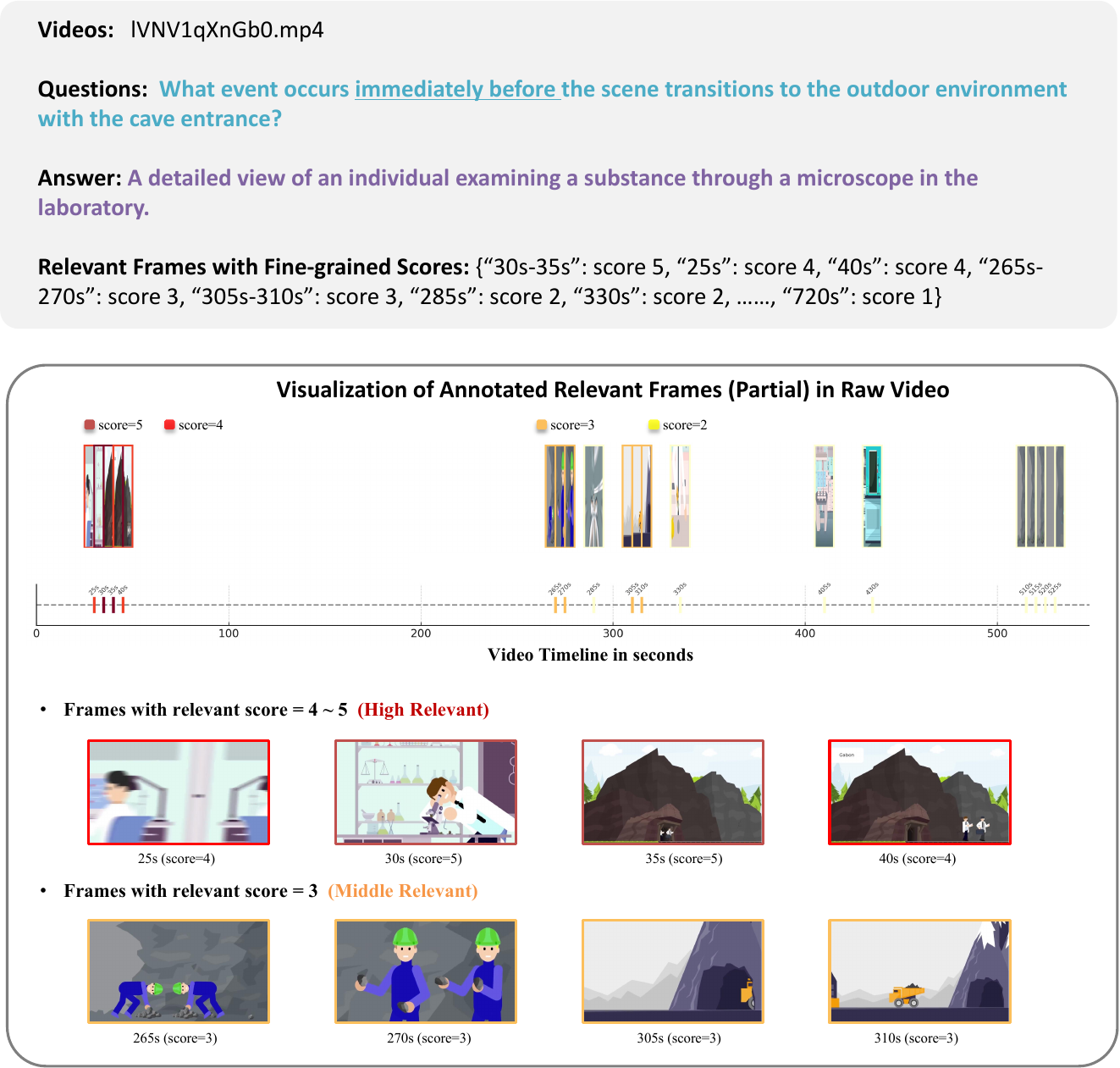}
    \caption{Visualization of annotated data sample from \dataset.}
    \label{fig:supp_data_case}
\end{figure*}

\end{document}